\pdfoutput=1

\documentclass[11pt]{article}

\usepackage[preprint]{acl}

\usepackage{times}
\usepackage{latexsym}

\usepackage[most]{tcolorbox}
\usepackage{amssymb}
\usepackage{hyperref}
\usepackage{url}
\usepackage{enumitem, tabularx}
\usepackage{graphicx}
\usepackage{wrapfig}
\usepackage{caption}
\usepackage{enumitem, tabularx}
\usepackage{multirow}
\usepackage{booktabs}
\usepackage{microtype}
\usepackage{algorithmic}
\usepackage{multicol}
\usepackage{float}
\usepackage[ruled,noline]{algorithm2e}
\newcounter{inlineenum}
\usepackage{colortbl}
\usepackage{subcaption}

\renewcommand{\theinlineenum}{\alph{inlineenum}}
\newenvironment{inlineenum}
  {\unskip\ignorespaces\setcounter{inlineenum}{0}%
   \renewcommand{\item}{\refstepcounter{inlineenum}{\textit{\theinlineenum})~}}}
  {\ignorespacesafterend}

\usepackage[T1]{fontenc}

\usepackage[utf8]{inputenc}

\usepackage{microtype}

\usepackage{inconsolata}

\title{ELAD: Explanation-Guided Large Language Models Active Distillation}

\author{Yifei Zhang, Bo Pan, Chen Ling, Yuntong Hu, Liang Zhao \\
         Emory University \\
         \texttt{yifei.zhang2@emory.edu}}
         
\begin{document}
\maketitle
\begin{abstract}
The deployment and application of Large Language Models (LLMs) is hindered by their memory inefficiency, computational demands, and the high costs of API inferences. Traditional distillation methods, which transfer the capabilities of LLMs to smaller models, often fail to determine whether the knowledge has been sufficiently transferred, potentially resulting in high costs or incomplete distillation. In this paper, we propose an Explanation-Guided LLMs Active Distillation (ELAD) framework that employs an active learning strategy to optimize the balance between annotation costs and model performance. To improve efficient sample selection, we introduce an explanation-guided sample selection method that identifies samples challenging its reasoning by exploiting uncertainties in explanation steps. Additionally, we present a customized LLM-annotated explanation revision technique where the teacher model detects and corrects flaws in the student model's reasoning. Our experiments across various reasoning datasets demonstrate that our framework significantly enhances the efficiency of LLM knowledge distillation.


\end{abstract}

\section{Introduction}
The advancement of Large Language Models (LLMs)~\cite{brown2020language, hoffmann2022training, thoppilan2022lamda, touvron2023llama} has significantly impacted natural language processing, showcasing excellent \textit{in-context learning} and \textit{complex reasoning} capabilities. 
Yet, the deployment of these models is hindered by their extensive parameter count, leading to significant computational demands and financial burdens. For instance, deploying LLMs with $100$-$200$ billion parameters would require a cluster of NVIDIA A100 GPUs, where each GPU costs $\$30,000$ in today's market. While cloud computing offers a solution, the costs associated with such services can quickly accumulate. Specifically, a cluster of A100 GPUs costs upwards of $\$25$ per hour, which may lead to a staggering $\$18,000$ monthly if operated non-stop\footnote {https://charshift.com/llm-true-cost/}. This financial barrier makes it impractical for many institutions and research labs to adopt LLMs widely, especially in resource-constrained environments \citep{bai2024beyond}, limiting their audience and application scope. Additionally, relying on API calls to access pre-trained LLMs, e.g., GPT-4, also presents its challenges, including high usage fees, inability to run models locally for customization or fine-tuning, potential data transmission issues, and privacy concerns \citep{yao2023survey}.

Recent research on LLMs \textit{knowledge distillation}~\cite{hinton2015distilling} enables smaller models to achieve performance similar to LLMs by transferring reasoning capabilities to them, making them more computationally efficient. \cite{tang2019distilling, wang2021want, arora2022ask} demonstrate the training of smaller models using pseudo-labels generated by LLMs, wherein LLMs act as ``teachers'' to supervise the \textit{fine-tuning} of these ``student'' models. Recent works~\cite{magister2022teaching, ho2022large, chan2022knife, li2023symbolic, hsieh2023distilling}, focus on multi-task fine-tuning of student models. They utilize chain-of-thought (CoT) reasoning~\cite{wei2022chain} to generate both explanations and final answers generated by LLMs as pseudo-labels to jointly supervise small model fine-tuning. However, a major issue with existing fully supervised learning methods is that they do not sense whether the knowledge has been sufficiently distilled into the small model. Insufficient distillation can lead to suboptimal performance of the small model, while excessive distillation may incur unnecessarily high costs.

To address this challenge, we take LLM as an agent that guides small language models toward progressive improvement. Throughout this process, the LLM (teacher) can sense the weaknesses of the small language model (student) and customize its teaching accordingly. Formally, we propose an \underline{E}xplanation-Guided \underline{L}LMs \underline{A}ctive \underline{D}istillation (ELAD) framework that significantly enhances active learning through the use of LLM explanations. In each iteration, the framework encompasses a student reasoning task and a teacher reasoning task: the student model identifies samples it struggles to predict accurately and reasonably; subsequently, for these selected samples, the teacher model reviews the student's explanations, correcting any erroneous reasoning.

However, first, it is nontrivial to tackle the student task. Current sample selection methods typically focus on finding samples with the wrong predictions~\cite{lewis1995sequential,ren2021survey,bansal2023large}, but even if the prediction is correct, the reasoning process can be wrong or flawed. The selection of samples with bad reasoning goes beyond it and is yet to be well explored, which requires the student model to faithfully self-inspect its step-by-step explanation of its prediction and locate the flaw. To address this, we propose a novel explanation-guided sample selection method that identifies the samples that trouble its reasoning by exploiting explanation stepwise uncertainties. Second, how the teacher senses and corrects the flaws in student model reasoning is also a challenging problem. Merely prompting teacher and student to generate their respective explanations separately and compare their difference is problematic because a prediction could be led by different reasoning processes and different explanations (e.g., \textit{Rashomon Effect}~\cite{roth2002rashomon}). We need the teacher model to check the student model's explanation, locate the problem within its reasoning, and correct it, which is not well explored. To accomplish this, we propose a customized LLM-annotated explanation revision technique. It entails sequentially prompting the LLM with the explanation from the small model and then asking it to assess whether the current step is reasonable or necessitates revision.

We evaluate our framework across six reasoning benchmarks, comparing it against existing sample selection methods for active learning and LLMs explanation and answer generation methods. Our findings indicate that the proposed framework notably enhances annotating efficiency.


We summarize our main contributions as follows: \begin{inlineenum}
\item An \textit{Explanation-Guided LLMs Active Distillation} framework that enhances active learning, guided by explanations from small models, during the distillation of LLMs to smaller models. \item An explanation-guided sample selection method that identifies the samples that trouble the reasoning of language model by exploiting explanation stepwise uncertainties. \item A customized LLM-annotated explanation revision technique that allows LLM to teach customized knowledge by guiding the LLM to pinpoint and correct inaccuracies in the reasoning steps of small models. \item Extensive experiments demonstrate that the proposed framework can improve annotating efficiency.
\end{inlineenum}  

\section{Related Work}
\subsection{LLMs Knowledge Distillation}
DistilBert~\cite{sanh2019distilbert} achieves efficient distillation of the BERT transformer into a student model with minimal performance loss. Tinybert~\cite{jiao2019tinybert} introduces a loss term for matching hidden states between teacher and student. Works like~\cite{magister2022teaching}, Fine-tune-CoT~\cite{ho2022large}, KNIFE~\cite{chan2022knife}, SCoTD~\cite{li2023symbolic}, and Distilling step-by-step~\cite{hsieh2023distilling} emphasize multi-task fine-tuning of student models using both CoT reasoning explanations and LLM-generated answers. Distilling step-by-step~\cite{hsieh2023distilling} specifically uses task-specific prefixes in prompts to tailor model responses. Li et al.~\cite{li2022explanations} investigate methods for generating explanations to aid student model learning. SOCRATIC CoT~\cite{shridhar2023distilling} decomposes problems into subproblems to guide reasoning, while SCOTT~\cite{wang2023scott} uses teacher-generated explanations for training on a counterfactual reasoning objective, promoting self-consistency.

\subsection{Efficient Annotating and Active Learning}
\cite{bansal2023large} introduced a method using model uncertainty~\cite{lewis1995sequential}, dataset density~\cite{ren2021survey}, and conditional informativeness~\cite{bansal2023large} for one-shot informative sample selection to enhance annotation efficiency. However, these one-shot approaches fail to sense the sufficiency of annotation. Addressing this, active learning, particularly the pool-based paradigm, has been recognized as a crucial technique for reducing annotation costs~\cite{krishnakumar2007active, ren2021survey}. For NLP tasks, the most common strategy is based on the entropy of predicted tokens for sampling~\cite{zhang2022survey}. Additionally,~\cite{yao2023beyond} proposed a data diversity-based active learning sampling strategy, leveraging explanation annotations. BSDETECTOR~\cite{chen2023quantifying} introduces an uncertainty quantification technique for black-box LLMs, focusing on consistency~\cite{wang2022self}.

\subsection{Explanation Generation for LLMs}
Extractive explanations~\cite{lei2016rationalizing, yu2019rethinking} focus on identifying key elements within the input that justify a prediction. However, they are limited in explaining complex reasoning tasks that require detailed natural language explanations (free-text explanation)~\cite{camburu2018snli,rajani2019explain}.~\cite{narang2020wt5} advanced this by training models to generate explanations post-prediction. Self-rationalization models, such as those discussed by~\cite{wiegreffe2020measuring}, aim to simultaneously predict labels and generate text-based explanations. STaR~\cite{zelikman2022star} generates explanations by augmenting ground truth answers as hints when predicted answers are incorrect.~\cite{wei2022chain} introduced CoT prompting, which uses demonstrations in LLM prompting to elicit intermediate reasoning steps for explanations.~\cite{kojima2022large} demonstrated the zero-shot reasoning capabilities of LLMs by employing prompts like ``Let’s think step by step'' to generate an explanation. Tree of Thought (ToT)~\cite{yao2023tree, long2023large} generates reasoning explanations by recursively decomposing complex questions into simpler sub-questions, solving them individually, and integrating their answers.

\section{Preliminary Study}
\label{sec:preliminary}

\subsection{LLMs Reasoning}
In the zero-shot CoT prompting scenario, prompting a question \(q\) to an LLM triggers the generation of the completion which consists of an answer \(a\) and reasoning path (explanation) \(r\), modeled as \((a, r) \sim P(a, r \mid q)\). This process unfolds in an auto-regressive manner, generating \(r\) before \(a\) and formalizing the conditional probability of the answer as \(P(a \mid q) = P(a \mid q, r) \times P(r \mid q)\), where \(P(a \mid q, r)\) represents the probability of \(a\) given both \(q\) and \(r\), and \(P(r \mid q)\) denotes the probability of \(r\) given \(q\). In few-shot scenarios, demonstration triplets \(\{(q_i^p, a_i^p, r_i^p)\}_{i=1}^m\) are included before \(q\) in the prompt, facilitating contextual guidance and reasoned answer generation, with \(m\) indicating the number of demonstrations in the prompt.

The CoT prompting approach facilitates sequential reasoning in LLMs, generating a series of reasoning steps \(r = \{s_1, s_2, \ldots, s_n\}\), where \(n\) is the total number of steps. Each step \(s_i\) contributes cumulatively to the reasoning explanation, culminating in the final answer. Specifically, the probability of generating the explanation \(r\) given the question \(q\), denoted \(P(r \mid q)\), is expressed as a product of conditional probabilities, representing the step-by-step reasoning:
\begin{equation}
P(r \mid q) = \prod\nolimits_{i=1}^{n} P(s_i \mid q, s_1, \ldots, s_{i-1})
\end{equation}


where each \(s_i\) is predicated on the question \(q\) and the preceding steps \(s_1, \ldots, s_{i-1}\).

\subsection{LLMs Knowledge Distillation}
LLMs Knowledge distillation entails a process wherein a large teacher language model $\mathcal{T}$ transfers its knowledge to a small student language model $\mathcal{S}$. In this framework, given an unlabeled dataset $\mathcal{U}$, the teacher model generates a pseudo-answer and a pseudo-explanation for each question $q$ in dataset $\mathcal{U}$. These outputs are represented as the answer-explanation pair (completion) $\left(\hat{a},\hat{r}\right)$. This generation process is modeled as:
\begin{equation}
\left(\hat{a},\hat{r}\right) \sim \mathcal{T}\left(a,r \mid q\right)
\end{equation}
The result is a collection of triplets $\{(q,\hat{a},\hat{r})\}^{\left|\mathcal{U}\right|}$ as the training dataset $\mathcal{D}$. Subsequently, the student model $\mathcal{S}$ is fine-tuned using $\mathcal{D},$ employing the standard language modeling loss, formulated as:
\begin{equation}
\max E_{(q, \hat{a}, \hat{r})} \sim_{\mathcal{D}} [\mathcal{S}(\hat{a}, \hat{r} \mid q)]
\end{equation}

\begin{figure*}[ht]
    \centering
    \includegraphics[width=\linewidth]{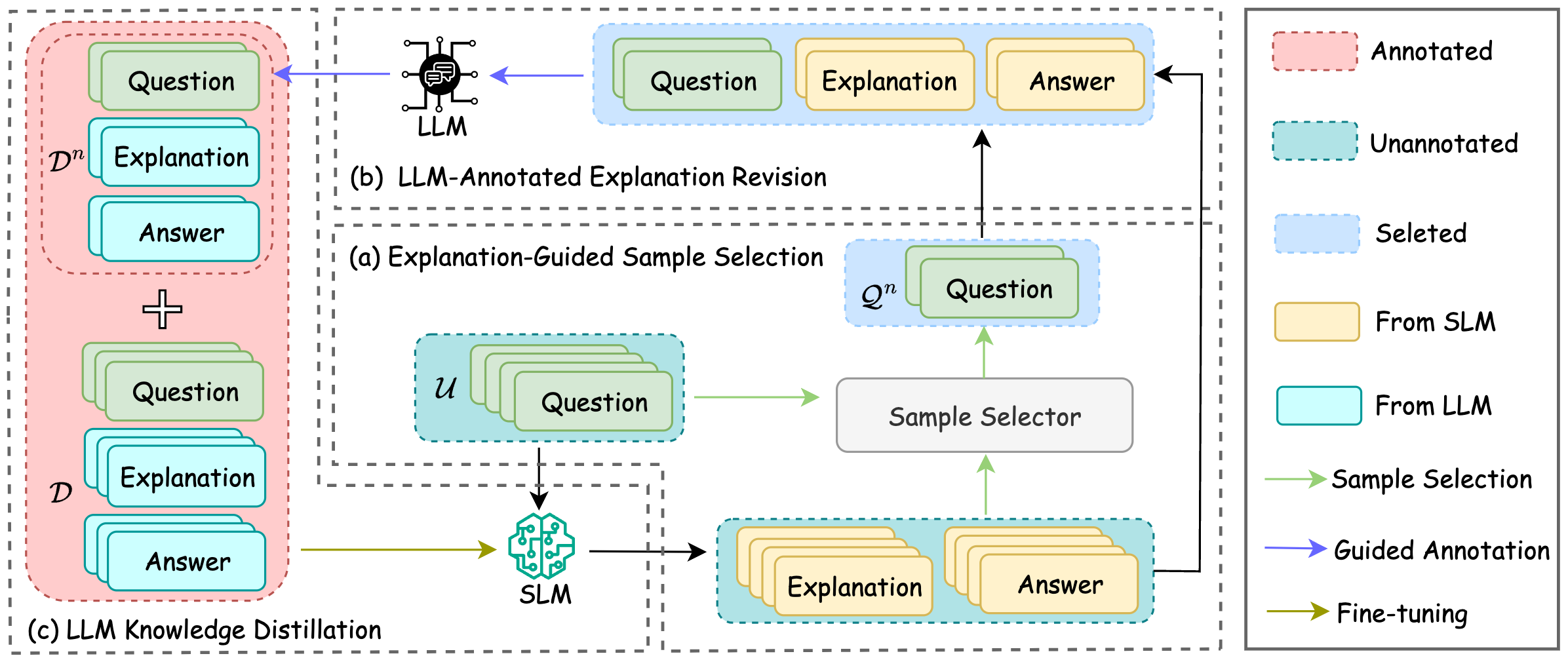}
    \caption{Overview of the Explanation-Guided LLM Active Distillation (ELAD) framework: (a) illustrates the Explanation-Guided Sample Selection method, (b) depicts the Customized LLM-Annotated Explanation Revision technique, and (c) showcases the LLM Knowledge Distillation (small model fine-tuning) process.}
    \label{fig:framework}
\end{figure*}

\section{Problem Setup}
In the context of knowledge distillation of LLM \(\mathcal{T}\), we address the problem of efficiently annotating an unlabeled question set \(\mathcal{U}\). This involves strategically selecting a subset \(\mathcal{Q}\) from \(\mathcal{U}\), where \(\mathcal{Q} \subset \mathcal{U}\) and \(|\mathcal{Q}| < |\mathcal{U}|\). Our study's primary goal is to enhance the performance of a smaller model \(\mathcal{S}\) through fine-tuning with a training dataset \(\mathcal{D} = \{(q_i,\hat{a}_i,\hat{r}_i)\}^{|\mathcal{Q}|}\), where the LLM annotates an answer \(\hat{a}\) and a corresponding explanation \(\hat{r}\) for each question \(q \in \mathcal{Q}\), forming \(\mathcal{D}\). The objective is to achieve a performance with \(\mathcal{S}\) that is comparable to that of \(\mathcal{T}\), while minimizing the size of the annotated dataset \(\mathcal{Q}\). This approach aims to maximize the efficiency of the small model and minimize the amount of annotated data required from the LLM.


\section{Methodology}
In this section, we first present the overview of our proposed \textit{Explanation-Guided LLMs Active Distillation} framework. We then proceed to present a novel explanation-guided sample selection method. Lastly, we present a customized LLM-annotated explanation revision method.

\subsection{Explanation-Guided LLMs Active Distillation Framework}
We propose a novel \textit{Explanation-Guided LLMs Active Distillation} framework to optimize the trade-off between sufficient distillation and annotation costs for LLM knowledge distillation tasks via an active learning strategy. Our overall framework is depicted in Figure~\ref{fig:framework}. We first collect an unlabeled dataset $\mathcal{U}$. At the \(n\)-th iteration of active learning, during the \textit{sample selection phase}, as depicted in Figure~\ref{fig:framework} (a), we employ the standard pool-based setting. The small model \(\mathcal{S}\) generates answers and explanations (completions) for all samples \(q \in \mathcal{U}\), resulting in a set \(\{(q_i, a_i, r_i)\}^{\left|\mathcal{U}\right|}\). Then, an \textit{explanation-guided sample selection} method \(f\) (details to be provided in Section~\ref{sec:sampling}) is adopted to select \(m\) samples with high \textit{uncertainty} in their generated answers and explanations from this set, forming the selected subset. This process can be represented as 
\begin{equation}
\label{eq:selector}
\mathcal{Q}^n = f(\{(q_i, a_i, r_i) \}^{\left|\mathcal{U}\right|}; m)
\end{equation}
We then create the batch \(\mathcal{B}^n\) comprising triples \(\{(q_i, a_i, r_i)\}^{m}\) for each \(q \in \mathcal{Q}^n\). Subsequently, we progress to the \textit{annotation phase} of active learning, as illustrated in Figure~\ref{fig:framework} (b). The customized LLM-annotated explanation revision function \(g\) (to be discussed in Section~\ref{sec:annotating}) annotates completions \((\hat{a},\hat{r})\) for the selected samples in \(\mathcal{Q}^n\) using the LLM \(\mathcal{T}\), guided by the completion \((a,r)\) generated by the small model, represented as 
\begin{equation}
\label{eq:annotating}
\mathcal{D}^n = g(\{(q_i, a_i, r_i)\}^{m}; \mathcal{T})
\end{equation}
For each \(q \in \mathcal{Q}^n\), this results in the dataset \(\mathcal{D}^n = \{(q_i, \hat{a}_i, \hat{r}_i)\}^{m}\) for small model fine-tuning. The datasets are updated by removing \(\mathcal{Q}^n\) from \(\mathcal{U}\) and adding \(\mathcal{D}^n\) to the cumulative training set \(\mathcal{D}\). Finally, in the \textit{model update phase} of active learning, depicted in Figure~\ref{fig:framework} (c), we fine-tune the small model \(\mathcal{S}\) on the training set \(\mathcal{D}\). This process is repeated until the LLM annotating (labeling) budget \(Bu\) is depleted or other stopping criteria are met (e.g., marginal improvement of the small model falls below a certain threshold). The overall algorithm is summarized in Algorithm~\ref{alg:ELAD}.

\begin{algorithm}
\caption{ELAD}
\small
\label{alg:ELAD}
\begin{algorithmic}[1]
  \REQUIRE $\mathcal{U}$, $\mathcal{D}$, $\mathcal{T}$, $\mathcal{S}$, annotating budget $Bu$, number of samples to select each iteration $m$
  \ENSURE Fine-tuned student model $\mathcal{S}$
  \WHILE{$Bu > 0$}
    \STATE Generate answers and explanations for samples \\ from $\mathcal{U}$ using small model $\mathcal{S}$
    \STATE Select $m$ most informative samples as $\mathcal{Q}^n$ using Equation~\ref{eq:selector} to form batch $\mathcal{B}^n$
    \STATE Annotate $\mathcal{B}^n$ using LLM $\mathcal{T}$ as per Equation~\ref{eq:annotating}, obtaining $\mathcal{D}^n$
    \STATE Update $\mathcal{D} \gets \mathcal{D} \cup \mathcal{D}^n$; $\mathcal{U} \gets \mathcal{U} \setminus \mathcal{Q}^n$
    \STATE Retrain $\mathcal{S}$ on updated $\mathcal{D}$
    \STATE $Bu \gets Bu - m$
  \ENDWHILE
  \STATE Perform final retraining of $\mathcal{S}$ on $\mathcal{D}$
\end{algorithmic}
\end{algorithm}

\subsection{Explanation-Guided Sample Selection}
\label{sec:sampling}
This section presents the \textit{explanation-guided sample selection} method to select samples with high \textit{uncertainty}. This \textit{uncertainty} stems from the complexity and instability inherent in the step-by-step reasoning process. We estimate it across two dimensions: 1) Intra-explanation uncertainty, which explores the uncertainty within individual steps of an explanation, and 2) Inter-explanation uncertainty, which examines the uncertainty across the aggregated answers from different reasoning paths.

\begin{figure}[ht]
    \centering
    \includegraphics[width=\linewidth]{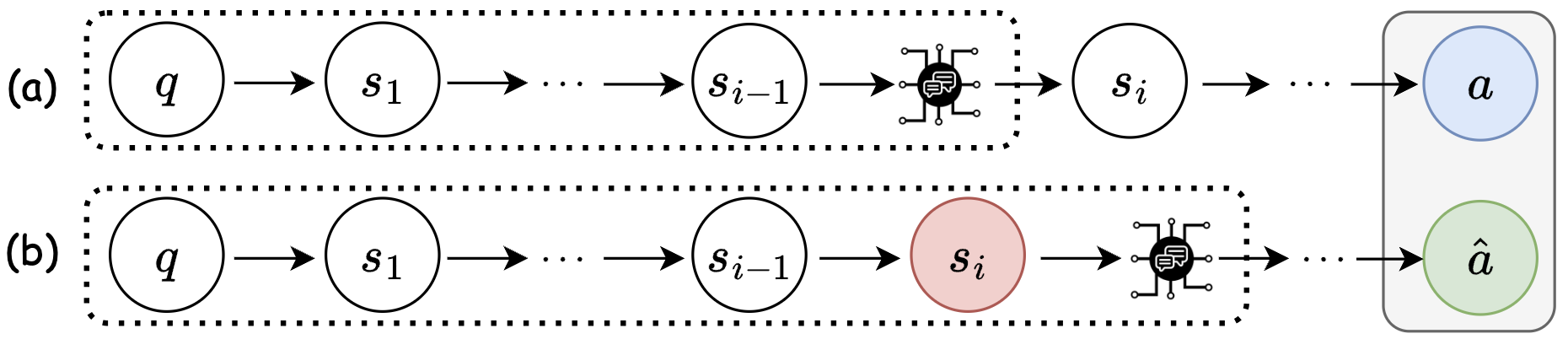}
    \caption{(a) illustrates reasoning not conditioned on the \(i\)-th reasoning step; (b) depicts reasoning conditioned on the \(i\)-th reasoning step.}
    \label{fig:stepaware}
\end{figure}

\noindent\textbf{Intra-explanation uncertainty} As stated in Section~\ref{sec:preliminary}, for an explanation \( r = \{s_1, s_2, \ldots, s_n\} \), each reasoning step \( s_i \) builds upon the question and all preceding steps and influences subsequent steps and the final answer. To address the challenge of estimating the uncertainty in the explanation generation resulting from step-by-step reasoning, we introduce a novel method for estimating intra-explanation uncertainty. This method utilizes a step-wise technique to evaluate the consistency of final answers, whether they are conditioned on specific reasoning steps or not. By comparing outcomes in both scenarios, we effectively measure the uncertainty associated with each step in the explanation. To be more specific, for the $i$-th reasoning step, the reasoning and answer before the $i$-th step can be written as: 
\begin{equation}
(a,s_{\ge i}) \sim \mathcal{S}(a,s_{\ge i} \mid q, s_{<n})
\end{equation}
Similarly, the reasoning process conditioned on the $i$-th reasoning step can be written as: 
\begin{equation}
(\hat{a},s_{>i}) \sim \mathcal{S}(\hat{a},s_{>i} \mid q, s_{<i}, s_{i})
\end{equation}
where $\hat{a}$ is the sampled answer conditioned on $i$-th reasoning step. The above two scenarios are illustrated in Figure~\ref{fig:stepaware}. We prompt the small model with \( (q, s_{<i}) \) and \( (q, s_{\leq i}) \) for each of the \(n\) steps to obtain the corresponding answers, resulting in a set \( \{(\hat{a}_i, a_i)\}_{i=1}^n \) that records the outcomes. The uncertainty of the explanation is then quantified by calculating the frequency of instances where predictions remain unchanged despite the removal of a reasoning step in the prompt as  
\begin{equation}
\label{eq:reasoning}
\mathcal{H}_{Reasoning}:=\frac{1}{n} \sum\nolimits_{i=1}^{n} \mathbb{I}\left(\hat{a}_i=a_i\right)
\end{equation}
where $\mathbb{I}\left(\hat{a}_i=a_i\right)$ is an indicator function that returns 1 (or 0) if the predicted answer is unchanged (or not). This intra-reasoning uncertainty score $\mathcal{H}_{Reasoning}$ measures the uncertainty of a single explanation.



\noindent\textbf{Inter-explanation uncertainty}
The answers and explanations generated by language models can exhibit diversity due to the randomness introduced by sampling temperature. To assess the uncertainty arising from this randomness, we propose a consistency-based method for evaluating inter-explanation uncertainty. For each question, we apply multiple path decoding strategy, prompting the model \(k\) times to generate \(k\) distinct reasoning paths. This process can produce different final answers, leading to \(N\) unique answer values, with each unique value occurring \(c_i\) times. We assess the consistency among these multiple final answers by calculating the frequency of occurrence \(c_i\) for every unique answer and subsequently computing the probability of each answer as \(p_i=\frac{c_i}{k}\). To quantify the uncertainty in the probability distribution of the output answers derived from multiple promptings, we utilize Shannon entropy, calculated as follows:
\begin{equation}
\label{eq:consistency}
\mathcal{H}_{Consistency}:=-\sum\nolimits_{i=1}^{N} p_i \log(p_i)
\end{equation}
This inter-explanation uncertainty score serves as an indicator of the model's reasoning uncertainty arising from different reasoning paths.

\noindent\textbf{Overall Uncertainty Estimation and Sample Selection}
Based on the two types of uncertainty illustrated in Equations~\ref{eq:reasoning} and \ref{eq:consistency}, we define the overall reasoning uncertainty $\mathcal{H}$ as:
\begin{equation}
\label{eq:H}
\mathcal{H}=\mathcal{H}_{Consistency} + \sum_{i=1}^{k}\nolimits \mathcal{H}_{Reasoning}^{(i)}
\end{equation}
For all samples in dataset \(\mathcal{U}\), we select $m$ samples with the highest uncertainty scores to form the selected subset \(\mathcal{Q}\). Based on the above, the Equation~\ref{eq:selector} is formalized as:
\begin{equation}
\mathcal{Q} = \underset{q \in \mathcal{U}}{\mathrm{argmax\text{-}m}} \, \mathcal{H}(q)
\end{equation}
where $\mathcal{H}(q)$ denotes the computed uncertainty score for a question $q$ using Equation~\ref{eq:H}.


\subsection{Customized LLM-Annotated Explanation Revision}
\label{sec:annotating}
After selecting the samples the small model troubles the reasoning, the next step is transferring knowledge from the LLM to the small model. This process involves using the LLM to generate pseudo-completion to fine-tune small model. To achieve this, we introduce a \textit{customized LLM-annotated explanation revision} technique. This approach leverages the advanced capabilities of the LLM to provide customized guidance to the small model by allowing the LLM to detect and correct flaws in the small model's reasoning.

\begin{figure}[ht]
    \centering
    \includegraphics[width=\linewidth]{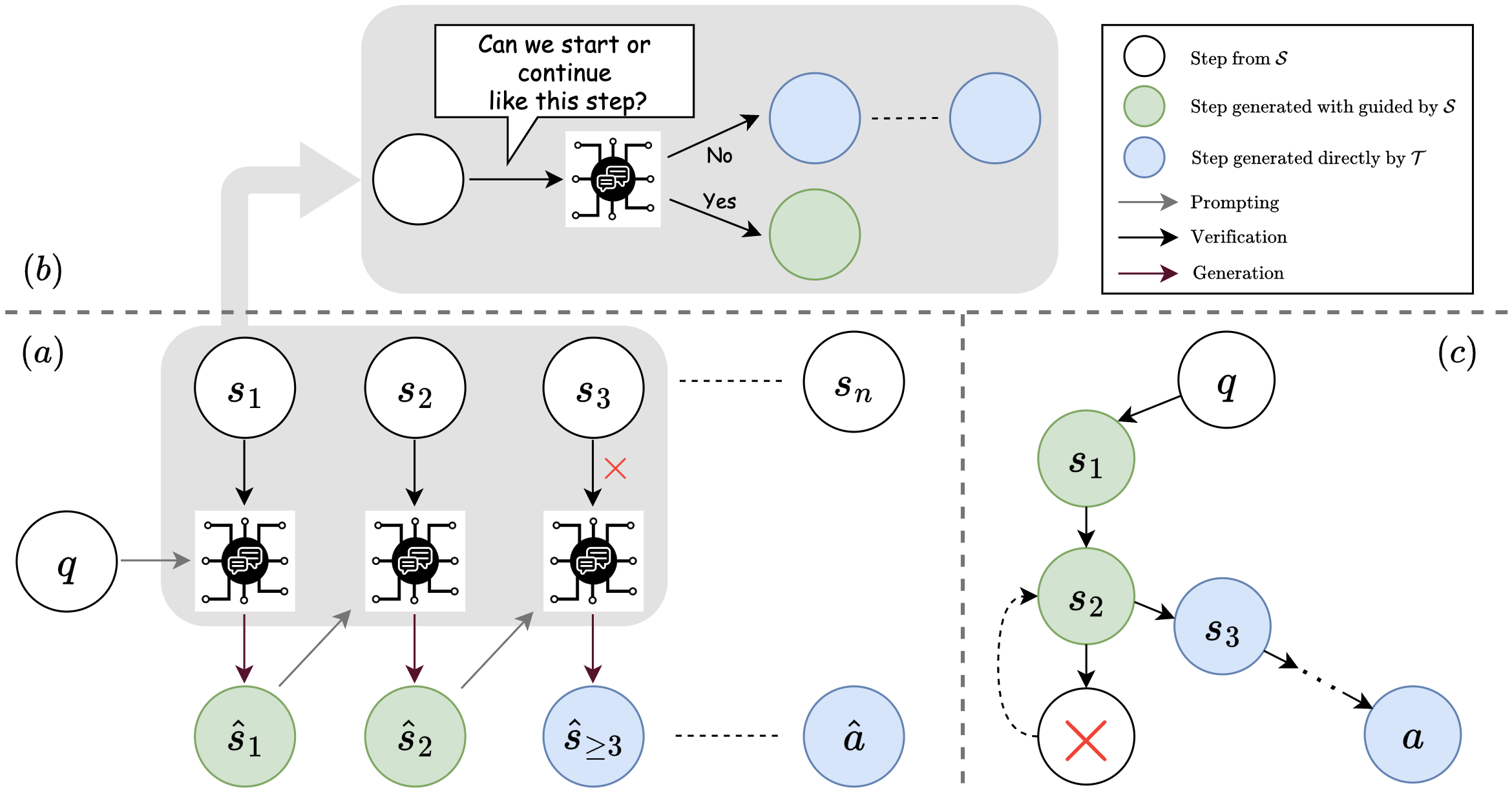}
    \caption{Customized LLM-Annotated Explanation Revision. (a) and (b) illustrate the process by which the LLM is prompted to revise the explanation and answer from the small model. (c) shows the DFS-based reasoning steps searching strategy.}
    \label{fig:alignment}
\end{figure}

As Figure~\ref{fig:alignment} (a) and (b) illustrate, our method prompts the LLM to annotate customized completion for the selected questions, conditioned on the output completion provided by the small model to make detection and possible revision (correction), as \((\hat{r},\hat{a}) \sim \mathcal{T}(\hat{r},\hat{a} \mid q,r,a)\). Specifically, we employ a Depth-First Search (DFS)-based strategy~\cite{yao2023tree}, where for each reasoning step generated by the small model, the LLM is prompted to perform verification to ascertain the validity of the current step. This verification process is iterative, continuing along the current reasoning path until it becomes infeasible to proceed further. At this point, the LLM is prompted to complete the reasoning process by generating the remaining steps and providing the final answer. As depicted in Figure~\ref{fig:alignment} (c), the process is represented as:
\(q \rightarrow s_1 \rightarrow \ldots \rightarrow s_{i-1} \rightarrow \times \rightarrow \hat{s}_i \rightarrow \ldots \rightarrow \hat{s}_n \rightarrow a\), denotes the $s_{i-1}$ reasoning step is infeasible. \( s_i \) are the steps generated by the small model, and \( \hat{s}_i \) are the steps generated (revised) by the LLM. 

Based on the above statement, the Equation~\ref{eq:annotating} can be formalized as the following process. We initiate the process with the question and prompt the LLM to determine if the current step \( s_i \) from the small model is valid for problem-solving. If the LLM's response is ``Yes'', we prompt it to generate its own reasoning step \( \hat{s}_i \) as \(\hat{s}_i \sim P(\hat{s}_i \mid q, s_i, \hat{s}_{<i}) \). This process is continued until the LLM responds ``No'' to the $i$-th step. At this point, we prompt the LLM to generate the remaining reasoning steps and the final answer as \((a, \hat{s}_{\geq i}) \sim P(a, \hat{s}_{\geq i} \mid q, \hat{s}_{<i}) \). The specifics of this verification and prompting process take the following form. Initially, we combine the first reasoning step from the student model \( s_1 \) with the question \( q \) to create the prompt: ``For question <$q$>, can we start with this step: <$s_1$>?'' If the LLM's answer is ``Yes'', we adopt the first reasoning step from the LLM \( \hat{s}_1 \) as the annotated step and proceed to the next reasoning step. For the \( i \)-th reasoning step from the student small model, we define the prompt as: ``Can we continue with this step: <$s_i$>?'' If the response is ``No,'' we then prompt the LLM with: ``What are the rest of the reasoning procedures and the answer?'' to generate the subsequent reasoning steps and the final answer. An example is shown below:

\begin{tcolorbox}[colback=blue!5!white,colframe=blue!50!black,title=\small{Customized LLM-Annotated Explanation Revision}]
\small 
\textbf{Prompt:} For question <$q$>. Let's think step by step. Can we start with this step: <$s_1$>? If yes, give me your step. If no, give me the rest steps and the final answer.\\
\textbf{Response:} Yes, we can start with that <$\hat{s}_1$>\\
\textbf{Prompt:} Can we continue with this step <\(s_2\)>?\\
\textbf{Response:} Yes, the second step is <$\hat{s}_2$>.\\
\textbf{Prompt:} Can we continue with this step <\(s_3\)>?\\
\textbf{Response:} No, we should proceed as <$\hat{s}_{\geq 3}$>, the final answer is <$\hat{a}$>.
\end{tcolorbox}

\begin{table*}[ht]
    \centering
    \small 
    \label{tab:exp}
    \resizebox{0.9\textwidth}{!}{%
    \begin{tabular}{lccccccc}
    \toprule       
    \multirow{2}{*}{Method} & \multirow{2}{*}{Annotating} & \multicolumn{2}{c}{\textit{Arithmetic}} & \multicolumn{2}{c}{\textit{NLI}} & \multicolumn{2}{c}{\textit{Commonsense}} \\
    \cmidrule(lr){3-4} \cmidrule(lr){5-6} \cmidrule(lr){7-8} & & GSM8K & AQuA & ANLI & e-SNLI & CommonSenseQA & StrategyQA \\
    \midrule
    \multicolumn{8}{c}{\textbf{Teacher:} \texttt{GPT-3.5-turbo}} \\
    \midrule
    Zero-shot-CoT & -- & 73.45 & 54.96 & 68.02 & 47.67 & 68.94 & 69.78 \\
    \midrule
    \multicolumn{8}{c}{\textbf{Student:} \texttt{LLaMA-2-7B}} \\
    \midrule
    Zero-shot-CoT & -- & 10.04 & 21.07 & 33.94 & 28.98 & 41.28 & 44.71  \\
    \midrule    
    \multicolumn{8}{c}{\textbf{Fine-Tuned Student}} \\
    \midrule 
    \multirow{2}{*}{Random} & \text{CoT Prompting} & 28.42 & 26.86 & 54.22 & 48.60 & 45.66 & 48.76 \\
    & \text{CLAER} & 30.31 & 27.05 & 57.12 & 48.56 & 48.54 & 50.89 \\
    \midrule
    \multirow{2}{*}{Maximum Entropy} & \text{CoT Prompting}& 27.58 & 27.67 & 52.56 & 47.98 & 46.35 & 49.03 \\
    & \text{CLAER} & 29.04 & 27.42 & 53.75 & 51.76 & 48.86 & 51.05 \\
    \midrule
    \multirow{2}{*}{Least Confidence} & \text{CoT Prompting}& 28.42 & 25.8 & 52.26 & 48.21 & 45.93 & 47.53 \\
    & \text{CLAER} & 28.68 & 27.19 & 53.63 & 48.65 & 48.52 & 51.23 \\
    \midrule
    \multirow{2}{*}{Disagreement} & \text{CoT Prompting}& 30.11 & 25.91 & 55.59 & 50.32 & 48.64 & 48.60 \\
    & \text{CLAER} & 31.49 & 27.23 & 58.71 & 54.32 & 52.46 & 53.81 \\
    \midrule
    \multirow{2}{*}{Self-Confidence} & \text{CoT Prompting}& 26.41 & 26.04 & 52.69 & 46.01 & 48.53 & 49.69 \\
    & \text{CLAER} & 27.95 & 25.57 & 54.32 & 49.21 & 49.03 & 52.44 \\   
    \midrule
    \multirow{2}{*}{EGSS} & \text{CoT Prompting}& 30.01 & 26.91 & 55.87 & 51.16 & 49.64 & 50.32 \\
    & \text{CLAER} & \cellcolor[HTML]{DAE8FC}32.72 & \cellcolor[HTML]{DAE8FC}28.43 & \cellcolor[HTML]{DAE8FC}58.02 & \cellcolor[HTML]{DAE8FC}54.44 & \cellcolor[HTML]{DAE8FC}53.53 & \cellcolor[HTML]{DAE8FC}55.63 \\ \midrule
    \bottomrule
    \end{tabular}}
    \caption{Performance of ELAD. Accuracy (\%) of Fine-tuned LLaMA-2 model with ELAD (EGSS and CLEAR) and with different baseline sample selection strategies and completion generation methods. We report results at 50\% annotation budget for all datasets in this table for comparison. \textcolor[HTML]{DAE8FC}{Blue} cells denote results of ELAD.}
\label{tab:main_results}
\end{table*}

\begin{figure*}[ht]
    \centering
    \begin{subfigure}[b]{0.3\textwidth} 
        \includegraphics[width=\textwidth]{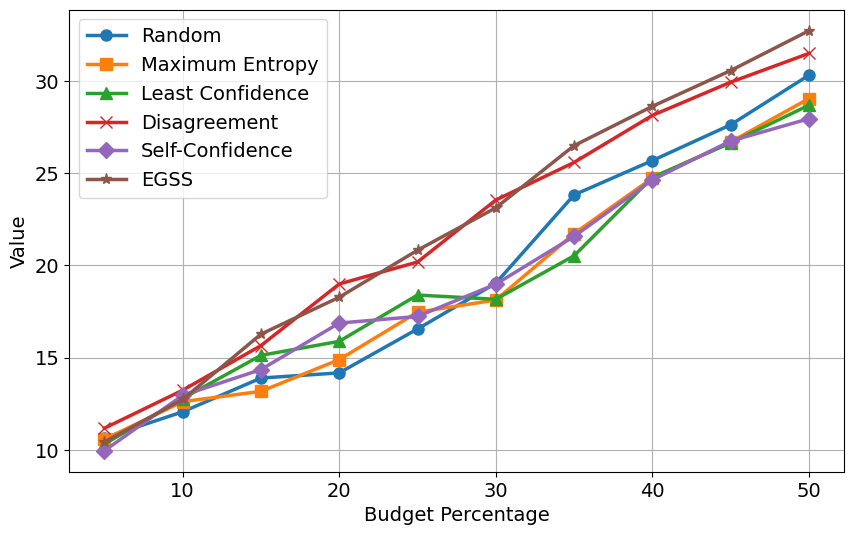}
        \caption{GSM8K}
    \end{subfigure}
    \begin{subfigure}[b]{0.3\textwidth} 
        \includegraphics[width=\textwidth]{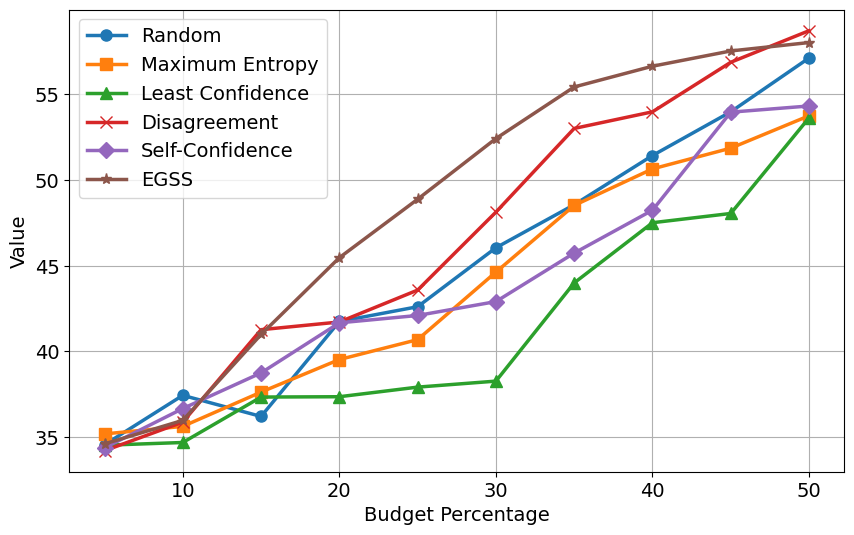}
        \caption{ANLI}
    \end{subfigure}
    \begin{subfigure}[b]{0.3\textwidth} 
        \includegraphics[width=\textwidth]{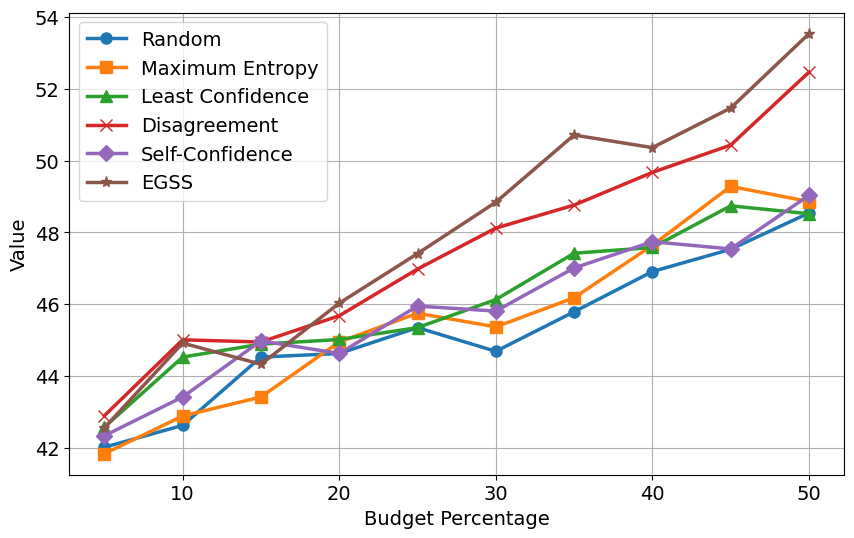}
        \caption{CommonSenseQA}
    \end{subfigure}
    
    \vfill 
    
    \begin{subfigure}[b]{0.3\textwidth} 
        \includegraphics[width=\textwidth]{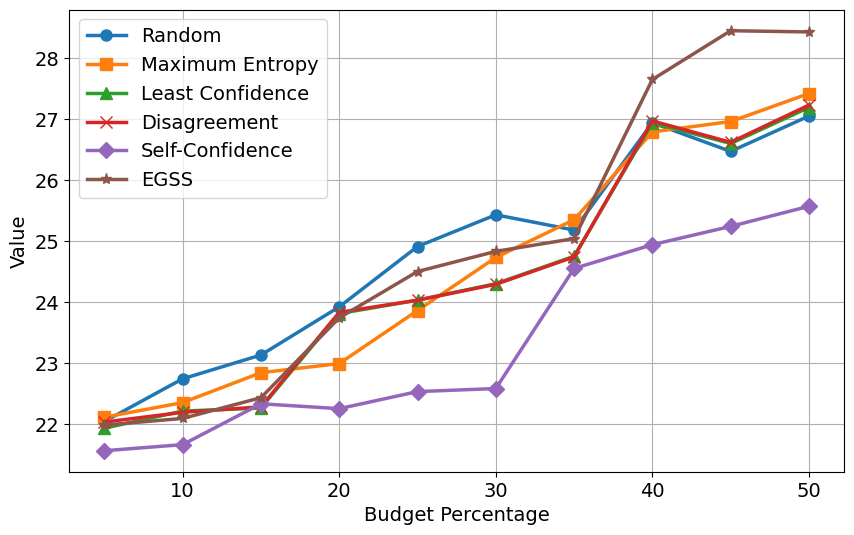}
        \caption{AQuA}
    \end{subfigure}
    \begin{subfigure}[b]{0.3\textwidth} 
        \includegraphics[width=\textwidth]{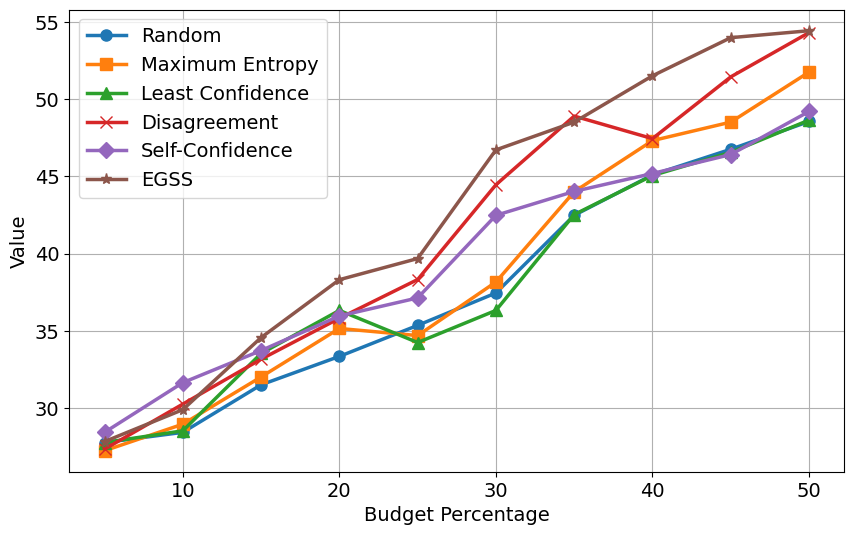}
        \caption{e-SNLI}
    \end{subfigure}
    \begin{subfigure}[b]{0.3\textwidth} 
        \includegraphics[width=\textwidth]{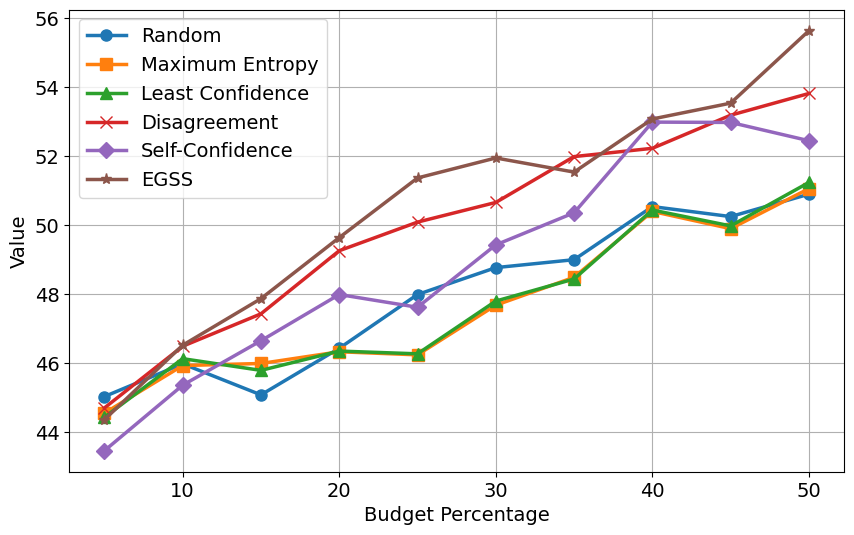}
        \caption{StrategyQA}
    \end{subfigure}
    \caption{Performance curves of different sample selection methods for active learning. The y-axis denotes the accuracy for the question-answering task, and the x-axis represents the percentage of samples annotated by the LLM for small model fine-tuning. In this case, 100\% denotes that all samples from the training set have been annotated. }
\label{fig:activelearning}
\end{figure*}

    

\section{Experiments}

\noindent\textbf{Datasets}
The experiments are conducted on six well-known benchmark datasets across 3 different reasoning tasks: GSM8K~\cite{cobbe2021training} and AQuA~\cite{geva2021did} for arithmetic reasoning tasks, ANLI~\cite{nie2019adversarial} and e-SNLI~\cite{camburu2018snli} for natural language inference (NLI) tasks, and StrategyQA~\cite{geva2021did} and CommonSenseQA~\cite{talmor2018commonsenseqa} for common sense reasoning task. Further details are provided in the Appendix~\ref{sec:appendix}.

\noindent\textbf{Evaluation Metric}
To assess question-answering performance for the above reasoning tasks, we calculate the \textit{accuracy} based on the final answers given by the student small model.

\noindent\textbf{Setup}
We use \texttt{GPT-3.5-turbo} as teacher via OpenAI API. We use \texttt{LLaMA-2-7B}~\cite{touvron2023llama} as our student model. Further implementation details are provided in the Appendix~\ref{sec:appendix}.

\noindent\textbf{Baseline Methods}
We compare the proposed ELAD framework with two different categories of baselines: 1) sample selection methods, and 2) completion generation methods. To be more specific, we provide a comparison of Explanation-Guided Sample Selection (EGSS) with five different sample selection methods: Random, Maximum Entropy~\cite{krishnakumar2007active}, Least Confidence~\cite{culotta2005reducing}, Disagreement (Vote Entropy)~\cite{engelson1996minimizing}, and Self-Confidence~\cite{kadavath2022language}. Further, we delineate the efficacy of our Customized LLM-Annotated Explanation Revision (CLEAR) method in contrast to the conventional vanilla annotating with zero-shot-CoT prompting~\cite{kojima2022large} method applied post-sample selection. We also include comparative results from student and teacher models, assessed without fine-tuning and using direct prompting to answer questions.

\subsection{Results and Analysis}
This section evaluates the reasoning performance of models using our proposed ELAD framework, comparing it with baseline methods. We highlight improvements in sample selection and completion generation. Performance trends from a 5\% to 50\% annotating budget are depicted in Figure~\ref{fig:activelearning}, illustrating the effectiveness of our EGSS method against other selection baselines. Furthermore, we detail reasoning performance at a 50\% annotating budget for our CR completion annotation method and the CoT prompting baseline in Table~\ref{tab:main_results}.

\noindent\textbf{Comparison with sample selection baselines} From Table~\ref{tab:main_results} we can observe that the EGSS method demonstrates significant performance improvements compared with traditional sample selection baselines for active learning. For arithmetic reasoning tasks (GSM8K and AQuA), EGSS with CLEAR exhibits a remarkable performance advantage. Specifically, it shows an increase of approximately 2.41\% and 1.38\% in accuracy over the next best-performing method for GSM8K and AQuA, respectively. In the context of natural language inference and commonsense reasoning tasks, such as ANLI, e-SNLI, CommonSenseQA, and StrategyQA, EGSS continues to set the benchmark. For instance, in the ANLI dataset, EGSS achieves a performance boost of nearly 3.27\% over the Least Confidence method with CLEAR. Similarly, for StrategyQA, EGSS demonstrates a substantial increase of 4.82\% in accuracy compared to the Disagreement strategy. From Figure~\ref{fig:activelearning}, It is evident that the proposed EGSS method effectively selects the most informative unlabeled questions, as evidenced by performance gains that align with increases in annotation budget. Initially, differences between EGSS and Disagreement strategies are minimal, likely due to the dominance of Inter-explanation uncertainty. However, as the annotation budget grows, EGSS significantly outperforms the Disagreement strategy, highlighting the crucial role of Intra-explanation uncertainty in identifying the most valuable samples for annotation.

\noindent\textbf{Evaluating customized LLM-annotated explanation revision method} Table~\ref{tab:main_results} showcases the Customized Revision technique's effectiveness over the baseline CoT Prompting across several tasks. In arithmetic tasks like GSM8K and AQuA, Customized Revision outperforms vanilla CoT Prompting annotation method by up to 2.71\% and 1.52\%, respectively, under the EGSS framework, highlighting its superior capability in refining reasoning skills. In NLI and Commonsense Reasoning tasks, such as ANLI and StrategyQA, Customized Revision demonstrates notable accuracy improvements of 2.15\% and 5.31\%, respectively. These results underline the method's strength in leveraging detailed explanations to enhance model understanding and performance significantly. 

\subsection{Ablation Studies}
In this section, we conduct an ablation study to investigate the importance of each component in the ELAD framework we propose, and the results are reported in Table~\ref{tab:abl}. The results reveal that the full proposed ELAD framework outperforms configurations lacking EGSS (w/o EGSS) and CLAER (w/o CLAER) across all tasks. Ours demonstrates a notable performance advantage, with improvements up to 2.41\% in arithmetic tasks, 6.88\% in NLI tasks, and 5.09\% in commonsense reasoning tasks over the "w/o EGSS" setup. This highlights the critical contributions of EGSS and CLAER to the framework's overall performance. The diminished performance in configurations without these components underscores their importance in enhancing model reasoning ability.

\begin{table}[ht]
\centering
\resizebox{\columnwidth}{!}{%
\begin{tabular}{@{}lcccccc@{}}
\toprule
\multirow{2}{*}{Setting} & \multicolumn{2}{c}{\textit{Arithmetic}} & \multicolumn{2}{c}{\textit{NLI}} & \multicolumn{2}{c}{\textit{Commonsense}} \\ 
\cmidrule(l){2-3} \cmidrule(l){4-5} \cmidrule(l){6-7} & GSM8K & AQuA & ANLI & e-SNLI & CommonSenseQA & StrategyQA    \\ \midrule
ELAD (Ours)  & 32.72 & 28.43 & 58.02 & 54.44 & 53.53 & 55.63 \\
w/o EGSS     & 30.31 & 27.05 & 57.12 & 48.56 & 48.54 & 50.89 \\
w/o CLAER    & 30.01 & 26.91 & 55.87 & 51.16 & 49.64 & 50.32 \\ \midrule
\bottomrule
\end{tabular}%
}
\caption{Ablation Study. We report the performance of our ELAD framework under different settings.}
\label{tab:abl}
\end{table}

\section{Conclusion}
This paper introduced the Explanation-Guided LLMs Active Distillation (ELAD) framework to address the challenges of deploying LLMs due to the high memory and computational demands. Our proposed framework achieves LLMs active distillation with explanation-guided sample selection and a customized LLM-annotated explanation revision. Extensive experiments on various reasoning datasets demonstrate the effectiveness of our approach in enhancing the distillation efficiency.

\section*{Limitations}
Our method, which utilizes LLMs as agents in active learning, is influenced by the design of prompts for the LLM, potentially affecting the quality of generated explanations and answers. Similarly, the prompt design for the small model can impact its reasoning abilities. Additionally, our approach requires submitting questions (data) to third-party services via APIs (e.g., OpenAI), posing a risk of data leakage. Additionally, due to budget constraints, we did not utilize the most recently released GPT-4.0 as the teacher model in our experiments. We plan to explore this in future research.

\section*{Ethical Considerations}
All datasets and models used in this study are open-source, and references to previous work are properly cited. For fine-tuning the small language model, we solely used triples generated by GPT-3.5 Turbo and LLaMA-2, both of which are publicly accessible. This work complies with ethical guidelines, and no ethical concerns have been identified.
\bibliography{custom}

\appendix
\section{Appendix}
\label{sec:appendix}

\subsection*{Datasets Details}

We provide more detailed descriptions of the datasets used in our experiments. We include a more detailed introduction and original sources released from the authors as follows:

\noindent \textbf{GSM8K (Grade School Math 8K)}~\cite{cobbe2021training}: A dataset containing approximately 8,000 math word problems designed for grade school students, testing a variety of mathematical skills in natural language. For more information, visit the \href{https://github.com/openai/grade-school-math}{GSM8K GitHub repository}.
    
\noindent \textbf{AQuA (Algebra Question Answering)}~\cite{geva2021did}: Features algebraic word problems with multiple-choice answers, aimed at evaluating algebraic problem-solving in AI systems. Available on \href{https://www.kaggle.com/bigquery/algebra-question-answering}{Kaggle}.
    
\noindent \textbf{ANLI (Adversarial Natural Language Inference)}~\cite{nie2019adversarial}: A dataset with natural language inference tasks, including adversarial examples, to test models' understanding of human language beyond existing NLI datasets. For more details, refer to the \href{https://github.com/facebookresearch/anli}{ANLI GitHub repository}.
    
\noindent \textbf{e-SNLI (Explainable Stanford Natural Language Inference)}~\cite{camburu2018snli}: Extends the SNLI dataset by providing human-annotated explanations for NLI decisions, assessing models on inference and explanation generation. Visit the \href{https://github.com/OanaMariaCamburu/e-SNLI}{e-SNLI GitHub repository} for more information.
    
\noindent \textbf{CommonSenseQA}~\cite{talmor2018commonsenseqa}: A question-answering dataset that focuses on commonsense reasoning, requiring an understanding of everyday concepts for correct answers. More details can be found on the \href{https://www.tau-nlp.org/commonsenseqa}{CommonSenseQA website}.
    
\noindent \textbf{StrategyQA}~\cite{geva2021did}: Tests models on strategic question answering, particularly on reasoning about implicit strategies for yes/no questions. Information is available on the \href{https://allenai.org/data/strategyqa}{AllenAI website}.
\begin{table}[h]
\resizebox{\columnwidth}{!}{%
\begin{tabular}{lcccc}
\toprule
Dataset & Task Type & \#Train & \#Validation & \#Test \\
\midrule 
GSM8K & Arithmetic & 7,473 & -- & 1,319 \\
AQuA & Arithmetic  & 10,000 & -- & 254 \\
ANLI & NLI & 16,946 & 1,000 & 1,000\\
e-SNLI & NLI & 549,367 & 9,842 & 9,824 \\
CommonSenseQA & Commonsense & 9,741 & 975 & 1,221 \\
StrategyQA & Commonsense & 1,603 & 490 & 687 \\
\bottomrule
\end{tabular}
}
\caption{Dataset statistics used in our experiments.}
\label{tab:datasets}
\end{table}

For each dataset where a validation set is not originally provided, we randomly subsample 10\% of the original training set to serve as a validation set. The dataset statistics are provided in Table~\ref{tab:datasets}.

\subsection*{Implementation Details}

The main experiments were conducted on a single NVIDIA GTX 3090 GPU with 24GB of memory. We utilize QLoRA4~\cite{dettmers2023qlora} by default to conduct parameter-efficient fine-tuning. We set the attention dimension as 64, the alpha parameter as 16, and the dropout rate as 0.1. For fine-tuning the \texttt{Llama-2-7b-chat-hf} model, we utilized the Huggingface libraries\footnote {https://huggingface.co/meta-llama/Llama-2-7b-chat-hf} and set the batch size to 16 and the learning rate to 1e-4 using the AdamW optimizer.

\subsection*{Examples for Customized LLM-Annotated Explanation Revision }
We present two examples of Customized LLM-Annotated Explanation Revision. Example \ref{tab:example1} illustrates a case where the small model makes incorrect reasoning in step 3. The explanation from the small model is sequentially prompted to LLMs for verification. At step 3, the LLM responds with ``No'' and proceeds to revise step 3 and generate the remaining steps and the final answer. Example \ref{tab:example2} provides an example where the small model errs in step 1 of its reasoning. The step 1 from the small model fails the LLM's verification, and LLM directly generates the entire completion.

\begin{table*}[ht]
\small
    \renewcommand{\arraystretch}{1.25} 
    \resizebox{\linewidth}{!}{
        \begin{tabular}{p{\textwidth}}
        \toprule 
        \textbf{Question:} Anne purchased a box of sweets that contains 15 packs, and each pack has 60 pieces. She kept two packs and gave the rest to her 10 friends equally. How many pieces of sweets did each of her friends receive? \\
        \textbf{Response from Small Model:} Let's think step by step. Step 1: Identify the information given in the problem. Number of packs of sweets Anne purchased: 15 Number of pieces in each pack: 60 Number of friends Anne gave the sweets to: 10. Step 2: Identify the unknown quantity. Number of pieces of sweets each friend received: x (where x is the unknown quantity). Step 3: Use the information given in the problem to find the unknown quantity. Since Anne kept two packs for herself, the total number of pieces she gave to her friends is 15 x 10 = 150. Therefore, the number of pieces each friend received is 150 / 10 = 15 pieces. So, each of Anne's 10 friends received 15 pieces of sweets from her. \\
        \hline
        \textbf{Prompt:} Anne purchased a box of sweets that contains 15 packs, and each pack has 60 pieces. She kept two packs and gave the rest to her 10 friends equally. How many pieces of sweets did each of her friends receive? Let's think step by step. Can we start with this step: Step 1: Identify the information given in the problem. Number of packs of sweets Anne purchased: 15 Number of pieces in each pack: 60 Number of friends Anne gave the sweets to: 10? Yes or no? If yes, give me your answer for this step. If no, give me the rest steps and final answer. \\
        \hline
        \textbf{Response:} \textcolor{green}{\checkmark} Yes, we can start with that step. Step 1: Identify the information given in the problem. Number of packs of sweets Anne purchased: 15. Number of pieces in each pack: 60. Number of friends Anne gave the sweets to: 10. \\
        \hline
        \textbf{Prompt:} Can we continue with this step Step 2: Identify the unknown quantity. Number of pieces of sweets each friend received: x (where x is the unknown quantity)? \\
        \hline
        \textbf{Response:} \textcolor{green}{\checkmark} Yes, we can continue with Step 2: Step 2: Identify the unknown quantity. Number of pieces of sweets each friend received: x (where x is the unknown quantity). \\
        \hline
        \textbf{Prompt:} Can we continue with Step 3: Use the information given in the problem to find the unknown quantity. Since Anne kept two packs for herself, the total number of pieces she gave to her friends is 15 x 10 = 150. Therefore, the number of pieces each friend received is 150 / 10 = 15 pieces.? \\
        \hline
        \textbf{Response:} \textcolor{red}{\(\times\)} No, we should proceed as Step 3: Use the information given in the problem to find the unknown quantity. Since Anne kept two packs for herself, the total number of pieces she gave to her friends is (15 packs - 2 packs) x 60 pieces/pack = 13 packs x 60 pieces/pack = 780 pieces. Therefore, the number of pieces each friend received is 780 pieces / 10 friends = 78 pieces. So, each of Anne's friends received 78 pieces of sweets. \\
        \bottomrule
        \end{tabular}
    }
\caption{Example that small model makes incorrect reasoning in its step 3}
\label{tab:example1}
\end{table*}

\begin{table*}[ht]
\small
    \renewcommand{\arraystretch}{1.25} 
    \resizebox{\linewidth}{!}{
        \begin{tabular}{p{\textwidth}}
        \toprule 
        \textbf{Question:} Rong has been saving 20 coins in his piggy bank every month. Neil has been saving 2/5 times more coins in his piggy bank per month than Rong. How many coins are they having ten years after they started their savings? \\
        \textbf{Response from Small Model:} Rong has been saving 20 coins in his piggy bank every month. Neil has been saving 2/5 times more coins in his piggy bank per month than Rong. How many coins are they having ten years after they started their savings? Let's think step by step. Rong has been saving 20 coins per month for 10 years, so he has saved a total of 20 x 10 = 200 coins. Neil has been saving 2/5 times more coins than Rong, so he has saved 2/5 x 20 = 40 coins per month. He has saved a total of 40 x 10 = 400 coins. Therefore, Rong has 200 coins and Neil has 400 coins after 10 years of saving." \\
        \hline
        \textbf{Prompt:} Rong has been saving 20 coins in his piggy bank every month. Neil has been saving 2/5 times more coins in his piggy bank per month than Rong. How many coins are they having ten years after they started their savings? Let's think step by step. Can we start with this step: Rong has been saving 20 coins per month for 10 years, so he has saved a total of 20 x 10 = 200 coins?  If yes, give me your answer on this step. If no, give me the rest steps and final answer. \\
        \hline
        \textbf{Response:} \textcolor{red}{\(\times\)} No, we cannot start with that step because Neil has been saving a different amount of coins per month than Rong. We need to calculate Neil's savings separately. Let's break it down step by step: Step 1: Calculate how many coins Rong has saved in 10 years. Rong saves 20 coins per month, so in a year, he saves 20 coins x 12 months = 240 coins. In 10 years, Rong will have saved 240 coins/year x 10 years = 2,400 coins. Step 2: Calculate how many coins Neil has saved in 10 years. Neil has been saving 2/5 times more coins than Rong per month. 2/5 of 20 coins = (2/5) * 20 = 8 coins. So, Neil saves an extra 8 coins per month compared to Rong. Neil's monthly savings = 20 coins (Rong's savings) + 8 coins (extra savings) = 28 coins per month. In a year, Neil saves 28 coins x 12 months = 336 coins. In 10 years, Neil will have saved 336 coins/year x 10 years = 3,360 coins. Step 3: Calculate the total number of coins they both have together. Rong saved 2,400 coins, and Neil saved 3,360 coins. So, the total number of coins they have saved together is 2,400 coins + 3,360 coins = 5,760 coins. So, after 10 years of saving, they will have a total of 5,760 coins. \\
        \bottomrule
        \end{tabular}
        }
\caption{Example that small model makes incorrect reasoning in its step 1}
\label{tab:example2}
\end{table*}

\end{document}